\documentclass{article}

\usepackage[utf8]{inputenc}
\usepackage{graphics}
\usepackage{multirow}

\usepackage[dvipsnames]{xcolor} 

\usepackage{graphicx}
\usepackage{amsfonts}
\usepackage{enumitem}
\usepackage[toc,page]{appendix}
\usepackage{listings}       
\usepackage{cprotect}

\begin{document}

\title{Language coverage and generalization \\ in RNN-based continuous sentence embeddings \\ for interacting agents}

\newcommand*\samethanks[1][\value{footnote}]{\footnotemark[#1]}

\author{Luca Celotti\thanks{These authors contributed equally.}\ , \hspace{1ex} Simon Brodeur\samethanks \ , \hspace{1ex} Jean Rouat\\
  NECOTIS, Département GEGI, 
  Université de Sherbrooke, 
Québec, Canada \\
  \texttt{ \{ Simon.Brodeur, Luca.Celotti, Jean.Rouat \}@USherbrooke.ca} \\
}
\date{October 2018}
\maketitle

\begin{abstract}
Continuous sentence embeddings using recurrent neural networks (RNNs), where variable-length sentences are encoded into fixed-dimensional vectors, are often the main building blocks of architectures applied to language tasks such as dialogue generation. While it is known that those embeddings are able to learn some structures of language (e.g. grammar) in a purely data-driven manner, there is very little work on the objective evaluation of their ability to cover the whole language space and to generalize to sentences outside the language bias of the training data.
Using a manually designed context-free grammar (CFG) to generate a large-scale dataset of sentences related to the content of realistic 3D indoor scenes, we evaluate the language coverage and generalization abilities of the most common continuous sentence embeddings based on RNNs. We also propose a new embedding method based on arithmetic coding, AriEL, that is not data-driven and that efficiently encodes in continuous space any sentence from the CFG.
We find that RNN-based embeddings 
underfit the training data and cover only a small subset of the language defined by the CFG. They also fail to learn the underlying CFG and generalize to unbiased sentences from that same CFG. We found that AriEL provides an insightful baseline.
\end{abstract}



\section{Introduction}


Several simulated 3D environments have emerged in the past two years as playgrounds for learning agents to solve language-based navigation \cite{Gupta2017,Parisotto2017,Devendra2018,Savinov2018} or general reasoning and manipulation tasks \cite{Das2017, Chaplot2017, Brodeur2017, Anderson2017,Hermann2017,Savva2017,Kolve2017, Yan2018, Wu2018} that require the agent to ground language related to the scenes.
Some of these environments \cite{Anderson2017, Savva2017, Kolve2017} aim at capturing the complexity of real-world indoor scenes. 
It is thus challenging for an agent to learn and efficiently represent all set of possible sentences related to the scene in a compact embedded space.
Recently, continuous sentence embeddings were successful in large-scale language tasks such as machine translation \cite{Sutskever2014} and goal-driven dialogues \cite{deVries2016,Strub2017}. They were also used for generative modeling of sentences \cite{Bowman2016} using sequence-to-sequence autoencoding (AE) \cite{Sutskever2014} and variational (VAE) \cite{Kingma2014} approaches. \cite{Kusner2017} augmented the variational approach with a context-free grammar (CFG) and was applied for the generation of arithmetic expressions . All these methods were shown to often produce grammatically-correct sentences, but language coverage
 was not evaluated.
It is not clear to which degree these embeddings are underfitting the data and represent only a fraction of the possible language space.
While the diversity of the output generated by VAE approaches can be measured by means of the entropy of the output and by the variety of unigrams and bigrams generated \cite{Bahuleyan2017}, this method doesn't scale well to the analysis of whole sentences.
Most of the related work \cite{Sutskever2014,deVries2016} is purely data-driven and have no access to the underlying grammar that generated the sentences. They are not able to quantify the ability of the agent to learn a given grammar, reconstruct and generate the full diversity of possible sentences. 
Our study is focused on the use of language embeddings based on recurrent neural networks and the evaluation of the language coverage and generalization ability they can provide. We therefore propose:
\begin{enumerate}[leftmargin=0.5cm] 
\item to measure the language coverage of several continuous sentence embedding approaches when trained from a large set of sentences generated by a known context-free grammar (CFG). An embedding that truly learned the underlying CFG should be able to reconstruct and generate any sentence that can be produced with that CFG.

\item to measure the generalization property of the continuous sentence embeddings when training on a biased dataset (reflecting real-life statistics on scenes in the SUNCG dataset \cite{Song2016}), but testing on a larger unbiased dataset from the same CFG (where objects have randomized attributes).
A latent space that truly learned the  CFG should perform equally well on both, biased and unbiased.

\item a continuous sentence embedding algorithm based on a multidimensional adaptation of arithmetic coding called AriEL. This method requires a CFG for encoding and decoding, and does not need 
learning. It provides an alternative and a reference that is not based on the neural network framework.
\end{enumerate}

\section{Optimal coding of context-free grammar in continuous spaces}

Arithmetic coding \cite{Rissanen1976,Rissanen1979,Witten1987} is one of the most commonly used algorithms in data compression to compact a sequence of symbols into a single real number of arbitrary precision (i.e. floating point value). Part of the family of entropy coding, it encodes frequently seen symbols with fewer bits than rare symbols. This makes the representation Shannon information optimal \cite{Shannon1948}.
We propose a continuous embedding algorithm based on a multidimensional adaptation of arithmetic coding, where sentences are encoded in $N_d$-dimensional space over the unit hypercube $[0, 1]^{N_d}$. This is illustrated in Figure \ref{fig:grammar_arithmetic_coding} for a 2D representation of a toy grammar (see Appendix \ref{sec:appendix_arithmetic_coding}).
The CFG is used to guide the partitioning of the unit hypercube based on which words are valid next, at any point in the sentence. The set of all possible sentences given by the CFG is thus encoded in a form very similar to a K-D tree, but where the partitioning can also depend on the probability of each word given its context. We name this method Arithmetic Embedding for Language (AriEL).

\begin{figure}[h!]
\centering
\includegraphics[width=0.6\linewidth]{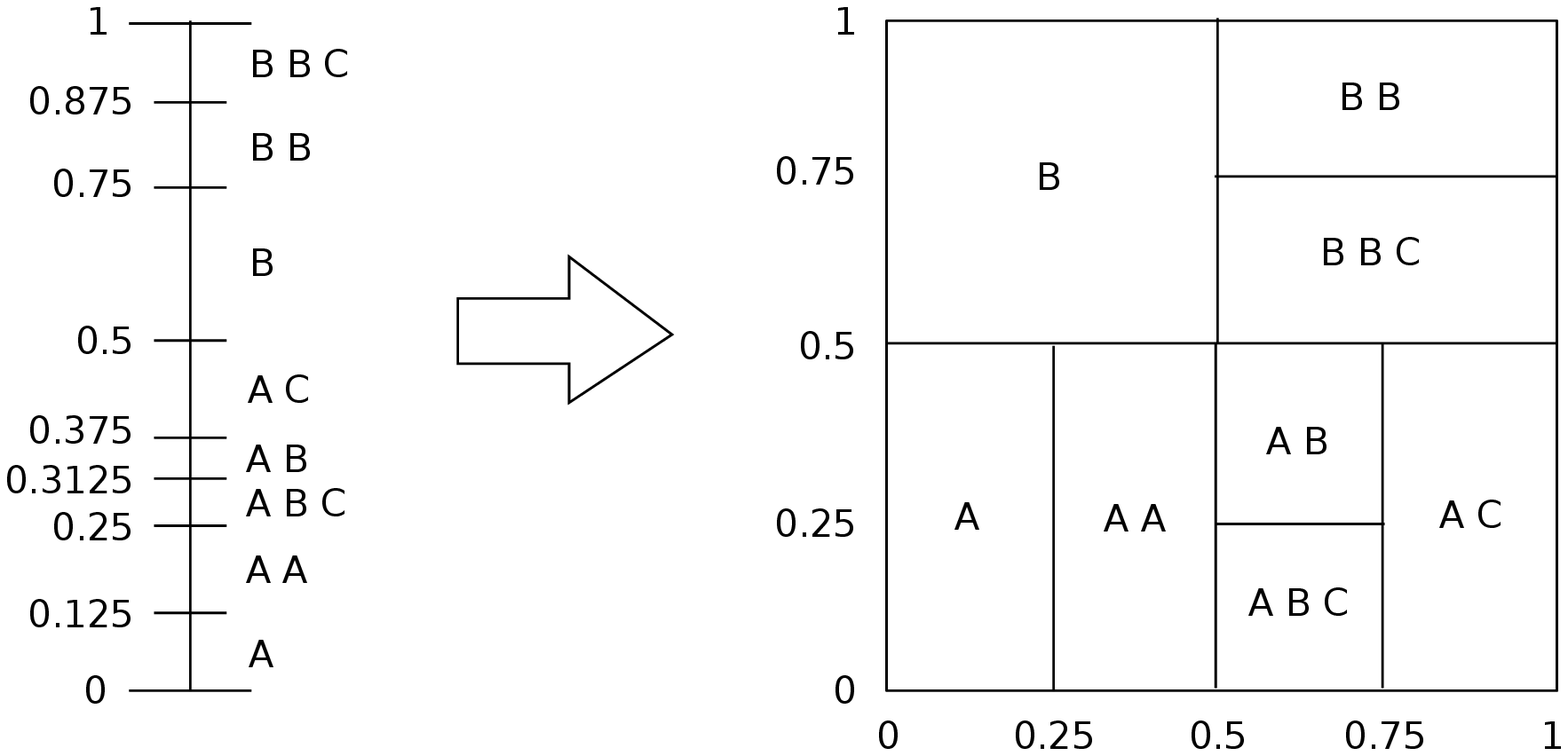}
\cprotect\caption{Continuous sentence embedding using arithmetic coding. In this example, the generating CFG is  \verb!S -> A | B | A A | A B | A C | B B | A B C | B B C! . Standard arithmetic coding (on the left) encodes any sequence of this CFG over a single dimension in the interval $[0, 1]$. The proposed multidimensional extension (on the right) allows to encode the CFG over higher dimensional spaces (here in 2D). For instance, the sequence "\verb!A B C!" could be encoded with AriEL as the vector $[0.625, 0.125]$. The simpler sentence "\verb!B!" could be AriEL encoded as $[0.25, 0.75]$, requiring less numerical precision. Long sentences cover smaller volumes of the partitioned space.}
\label{fig:grammar_arithmetic_coding}
\end{figure}

\section{Methodology}

\begin{figure}[h!]
\centering
\includegraphics[width=0.7\linewidth]{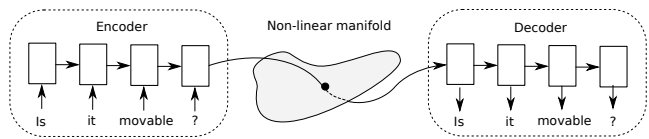}
\caption{Continuous sentence embedding using recurrent neural networks (RNNs), known as a sequence to sequence autoencoder \cite{Sutskever2014}. The input sentence (e.g. "is it movable ?") is fed to the RNN-based encoder, which sequentially accumulates information about individual words in its internal state. In the end of the sentence, this internal state represents a vector in the non-linear manifold where the complete sentence is embedded. A similar RNN-based decoder converts back the embedded vector into a sentence. 
In this framework, language coverage can be evaluated from two perspectives: (1) by encoding sentences from a dataset and looking at the reconstructions, or (2) by randomly sampling the non-linear manifold and looking at the generated sentences.}
\label{fig:seq2seq}
\end{figure}

\subsection{Context and experimental conditions}
We consider the family of approaches that maps variable length discrete spaces to fixed length continuous spaces, such as sequence to sequence autoencoders \cite{Sutskever2014} and their variational version \cite{Bowman2016}.
We stack two RNN layers with GRU units \cite{Cho2014} both, at the encoder and at the decoder to increase the representational capabilities \cite{Pascanu2014}. 
The last encoder layer has 
either $N_d = 16$ units or $N_d = 512$ for all methods. The output of the last encoder has a \emph{tanh} activation, to constraint the volume of the latent space and ease its sampling during evaluation.
The output of the decoder is a softmax distribution over the entire vocabulary. 
During testing, the output of the RNN is fed back to the unit. We used greedy decoding for all methods, but also allowed to use a language model (LM) based on the CFG during decoding. The language model was implemented by masking invalid words at each step during decoding (i.e. weighting the softmax distribution), from the set of next possible words that can be computed with the CFG, producing only grammatically correct sentences. The procedure is parallel to the one proposed in the Grammar VAE \cite{Kusner2017} to generate valid chemical structures.


\subsection{Dataset: grammar and vocabulary}

To create sentences that are \emph{biased} to the scenes (specific to the environment of the agent), we used the SUNCG large-scale dataset of 3D indoor scenes \cite{Song2016}. It provides 45k scenes and over 2500 objects with distinct properties (e.g. color, shape, texture). Questions about objects in the scenes are generated with a context-free grammar (CFG) (see Appendix \ref{sec:appendix_cfg}). The vocabulary consists of 840 words. 1M unique \emph{biased} sentences have been generated with the CFG. Of those, 10k sentences were exclusively used as the test set.
Another set of 10k  \emph{unbiased} sentences (not specific to the agent's environment) was also created with the same CFG to be used as another test set.
These sentences are not constrained by the SUNCG scenes.
While these \emph{unbiased} sentences are still grammatically correct (e.g. "Is it the wooden toilet in the kitchen ?"), they do not correspond to realistic situations.


\subsection{Objective evaluations}



\paragraph{Language coverage evaluation using generation (sampling) method}
It is evaluated by sampling the latent space of each embedding and retrieving the resulting sentences after the decoder. We sampled 10k sentences and applied those four measures:
\emph{i)} \textit{Grammar coverage} as the ratio of grammar rules (e.g. single adjective, multiple adjectives) that could be parsed in the sampled sentences;
\emph{ii)} \textit{Vocabulary coverage} as the ratio of words in the vocabulary that appeared in the sampled sentences;
\emph{iii)} \textit{Uniqueness} as a ratio of unique sampled sentences;
 and \emph{iv)} \textit{Validity} as a ratio of valid sampled sentences, meaning unique and grammatically correct. 

\paragraph{Language coverage evaluation using reconstruction method}
It is evaluated by encoding the 10k \emph{biased} sentences from the test set and looking at the reconstructions with the following objective criteria:
 \emph{i)} \textit{Reconstruction accuracy} as a ratio of correctly reconstructed sentences (i.e. all words must match);
\emph{ii)} \textit{Grammar accuracy} as a ratio of grammatically correct reconstructed sentences (i.e. can be parsed by the CFG);
 and \emph{iii)} \textit{Semantic accuracy} as a ratio of semantically correct reconstructed sentences. For instance, the sentences "is it blue and red ?" and "is it red and blue ?" are considered semantically identical.


\paragraph{Evaluation of generalization}
It was evaluated using the 10k \emph{unbiased} sentences while the embeddings were trained on the \emph{biased} training set. The \textit{reconstruction accuracy} of the \emph{unbiased} test set is computed and compared with the same metric on the \emph{biased} test set. It allows us to measure how well the latent space can generalize to grammatically correct (but albeit unusual) sentences outside the language bias. 

\section{Discussion and results}

Language coverage was evaluated for all embeddings using both generation (sampling) and reconstruction methods. The results are shown in Table \ref{tab:data}.
AE with LM and a latent dimension of 16, generates more valid sentences (unique and grammatical), 65\%, against the 39.7\% achieved by AriEL, which might be of interest for interactive agents. An AE without LM is able to produce many unique sentences, but mostly grammatical. Remarkably AE with LM was able to produce sentences that cover all the grammar rules. 
Both AE methods collapse in all but one measure, as we move from 16 to 512 units, suggesting overfitting.  VAE seems to improve with the latent size, but its overall performance remains very low. Both VAE methods have overlapping behaviors and LM gives no significant advantage.

Language coverage with the reconstruction method shows in Table \ref{tab:data} that AriEL is able to reconstruct any grammatically correct sentence. 
Interestingly having a language model at the output of the neural networks does not provide an advantage. The reconstruction seems to be always almost grammatically perfect, even if it does not coincide with the initial sentence. It is important to stress that VAE often learns to generate only one or few grammatically correct sentences independently of where the sampling is done in the latent space. VAE underperforms or matches AE based models.

The generalization abilities of the embeddings are shown on the last column of Table \ref{tab:data}. The large vocabulary, complex grammar, and the limits imposed in the latent space (small $N_d$ and \textit{tanh}), made it impossible for AE and VAE to achieve good accuracy. Removing some of these constraints gives better performance, primarily by removing the \textit{tanh} that was envisioned to allow for sampling from the latent space.
AE achieves 46.1\% over biased and 3.5\% with unbiased, both quite poor. VAE was incapable of learning the task at all. LM did not provide any benefit. The results for a 512 dimensional latent space are analogous or worse. AriEL achieves as expected perfect reconstruction.

\begin{table}
\resizebox{\columnwidth}{!}{
\begin{tabular}{@{\extracolsep{4pt}}crcccccccc@{}}
& &\multicolumn{4}{c}{Generation} &  \multicolumn{3}{c}{Reconstruction} & \multicolumn{1}{c}{Generalization} \\
 \cline{3-6} \cline{7-9} \cline{10-10} \\
&\textbf{model} & \shortstack{grammar \\ coverage} & \shortstack{vocabulary \\ coverage} &  \shortstack{validity} & uniqueness & \shortstack{semantic \\ accuracy} & \shortstack{grammar \\ accuracy} & \shortstack{reconstruction \\ accuracy \\ biased} & \shortstack{reconstruction \\ accuracy \\ unbiased} \\
\vspace{-.2cm} \\
\cline{2-10}
\vspace{.02cm} \\
\multirow{5}{*}{\rotatebox[origin=c]{0}{$N_d = 16$} $\left.
\begin{tabular}{l} \\ \\ \\ \\ \\ \end{tabular}
\right\{$} & \textbf{AriEL} & \textbf{100.0\%} & \textbf{57.0\%} & 39.7\% & 39.7\% & \textbf{100.0\%} & \textbf{100.0\%} & \textbf{100.0\%} & \textbf{100.0\%} \\
& \textbf{AE} & 71.4\% & 31.4\% & 16.8\% & \textbf{ 91.5\%} & 56.5\% & 97.7\% & 46.1\% & 3.5\% \\
& \textbf{AE-LM} & \textbf{100.0\%} & 33.8\% & \textbf{65.0}\% & 65.0\% & 56.6\% & \textbf{100.0\%} & 46.1\% & 3.5\% \\
& \textbf{VAE} & 28.6\% & 1.2\% & 0.0\% & 0.0\% & 0.0\% & \textbf{100.0\%} & 0.0\% & 0.0\% \\
& \textbf{VAE-LM} & 28.6\% & 1.2\% & 0.0\% & 0.0\% & 0.0\% & \textbf{100.0\%} & 0.0\% & 0.0\% \\
\vspace{.02cm} \\
\multirow{5}{*}{$N_d = 512$ $\left.\begin{tabular}{l} \\ \\ \\ \\ \\ \end{tabular}\right\{$} 
& \textbf{AriEL} & \textbf{100.0\%} & 53.1\% & 39.8\% & 39.8\% & \textbf{100.0\%} & \textbf{100.0\%} & \textbf{100.0\%} & \textbf{100.0\%} \\
& \textbf{AE} & 71.4\% & 39.5\% & 4.4\% & 75.3\% & 34.1\% & 98.6\% & 27.5\% & 3.5\% \\
& \textbf{AE-LM} & 85.7\% & 32.3\% & 29.0\% & 29.0\% & 34.1\% & \textbf{100.0\%} & 27.5\% & 3.5\% \\
& \textbf{VAE} & 42.9\% & 2.0\% & 0.0\% & 0.0\% & 0.0\% & \textbf{100.0\%} & 0.0\% & 0.0\% \\
& \textbf{VAE-LM} & 42.9\% & 2.1\% & 0.0\% & 0.0\% & 0.0\% & \textbf{100.0\%} & 0.0\% & 0.0\% \\
\end{tabular}
}
\vspace{.5cm}
\caption{\textbf{Evaluation of continuous sentence embeddings.} Complete results for the different methods and the different proposed measures, for varying dimensionality $N_d$ of the latent space.}
\label{tab:data}
\end{table}


\section{Conclusion and Future Work}

In this work, we used a manually designed context-free grammar (CFG) to generate our own large-scale dataset of sentences related to the content of realistic 3D indoor scenes. We found that RNNs-based continuous sentence embeddings largely underfit the training data and only cover a small subset of the possible language space. They also fail to learn the underlying CFG and generalize to unbiased sentences from that same CFG. We proposed a new continuous sentence embedding method based on a multidimensional extension of arithmetic coding, AriEL.
One current shortcoming of AriEL is generating a large diversity of unique sentences through stochastic sampling in the latent space. 
We conducted preliminary experiments (results not shown) that suggest AriEL might still provide a convenient embedded space to be used as a continuous action space for reinforcement learning dialogue tasks. The relation between coding of a CFG with AriEL and how RNN-based embeddings cover the large diversity of language will be studied in more depth.

\subsubsection*{Acknowledgments}
The authors would like to thank the ERA-NET (CHIST-ERA) and FRQNT organizations for funding this research as part of the European IGLU project. NVIDIA Corporation supported this research with the donation of a Titan X and Tesla K40.

\clearpage


\bibliographystyle{IEEEtran}

\begin{thebibliography}{30}

\bibitem{Gupta2017}
S.~Gupta, J.~Davidson, S.~Levine, R.~Sukthankar, and J.~Malik, ``Cognitive
  mapping and planning for visual navigation,'' in \emph{2017 IEEE Conference
  on Computer Vision and Pattern Recognition (CVPR)}, July 2017, pp.
  7272--7281.

\bibitem{Parisotto2017}
E.~Parisotto and R.~Salakhutdinov, ``Neural map: Structured memory for deep
  reinforcement learning,'' in \emph{International Conference on Learning
  Representations}, 2018.

\bibitem{Devendra2018}
D.~S. Chaplot, E.~Parisotto, and R.~Salakhutdinov, ``Active neural
  localization,'' in \emph{International Conference on Learning
  Representations}, 2018.

\bibitem{Savinov2018}
N.~Savinov, A.~Dosovitskiy, and V.~Koltun, ``Semi-parametric topological memory
  for navigation,'' in \emph{International Conference on Learning
  Representations}, 2018.

\bibitem{Das2017}
A.~Das, S.~Datta, G.~Gkioxari, S.~Lee, D.~Parikh, and D.~Batra, ``{E}mbodied
  {Q}uestion {A}nswering,'' in \emph{Proceedings of the IEEE Conference on
  Computer Vision and Pattern Recognition (CVPR)}, 2018.

\bibitem{Chaplot2017}
D.~S. Chaplot, K.~M. Sathyendra, R.~K. Pasumarthi, D.~Rajagopal, and
  R.~Salakhutdinov, ``Gated-attention architectures for task-oriented language
  grounding,'' in \emph{Proceedings of the Thirty-Second {AAAI} Conference on
  Artificial Intelligence, (AAAI-18), the 30th innovative Applications of
  Artificial Intelligence (IAAI-18), and the 8th {AAAI} Symposium on
  Educational Advances in Artificial Intelligence (EAAI-18), New Orleans,
  Louisiana, USA, February 2-7, 2018}, 2018, pp. 2819--2826.

\bibitem{Brodeur2017}
S.~Brodeur, E.~Perez, A.~Anand, F.~Golemo, L.~Celotti, F.~Strub, J.~Rouat,
  H.~Larochelle, and A.~Courville, ``{HoME: a Household Multimodal
  Environment},'' in \emph{{NIPS 2017's Visually-Grounded Interaction and
  Language Workshop}}, Long Beach, United States, Dec. 2017.

\bibitem{Anderson2017}
P.~Anderson, Q.~Wu, D.~Teney, J.~Bruce, M.~Johnson, N.~Sünderhauf, I.~Reid,
  S.~Gould, and A.~van~den Hengel, ``Vision-and-language navigation:
  Interpreting visually-grounded navigation instructions in real
  environments,'' in \emph{The IEEE Conference on Computer Vision and Pattern
  Recognition (CVPR)}, June 2018.

\bibitem{Hermann2017}
K.~M. {Hermann}, F.~{Hill}, S.~{Green}, F.~{Wang}, R.~{Faulkner}, H.~{Soyer},
  D.~{Szepesvari}, W.~M. {Czarnecki}, M.~{Jaderberg}, D.~{Teplyashin},
  M.~{Wainwright}, C.~{Apps}, D.~{Hassabis}, and P.~{Blunsom}, ``{Grounded
  Language Learning in a Simulated 3D World},'' \emph{ArXiv e-prints}, Jun.
  2017.

\bibitem{Savva2017}
M.~Savva, A.~X. Chang, A.~Dosovitskiy, T.~Funkhouser, and V.~Koltun, ``{MINOS:
  Multimodal Indoor Simulator for Navigation in Complex Environments},''
  \emph{ArXiv e-prints}, 2017.

\bibitem{Kolve2017}
E.~Kolve, R.~Mottaghi, D.~Gordon, Y.~Zhu, A.~Gupta, and A.~Farhadi,
  ``{AI2-THOR: An Interactive 3D Environment for Visual AI},'' \emph{ArXiv
  e-prints}, 2017.

\bibitem{Yan2018}
C.~Yan, D.~Misra, A.~Bennnett, A.~Walsman, Y.~Bisk, and Y.~Artzi, ``{CHALET:
  Cornell House Agent Learning Environment},'' \emph{ArXiv e-prints}, 2018.

\bibitem{Wu2018}
Y.~Wu, Y.~Wu, G.~Gkioxari, and Y.~Tian, ``{Building Generalizable Agents with a
  Realistic and Rich 3D Environment},'' \emph{ArXiv e-prints}, 2018.

\bibitem{Sutskever2014}
I.~Sutskever, O.~Vinyals, and Q.~V. Le, ``Sequence to sequence learning with
  neural networks,'' in \emph{Proceedings of the 27th International Conference
  on Neural Information Processing Systems - Volume 2}, ser. NIPS'14, 2014, pp.
  3104--3112.

\bibitem{deVries2016}
H.~De~Vries, F.~Strub, S.~Chandar, O.~Pietquin, H.~Larochelle, and A.~C.
  Courville, ``Guesswhat?! visual object discovery through multi-modal
  dialogue.'' in \emph{Conference on Computer Vision and Pattern Recognition
  (CPVR)}, vol.~1, no.~2, 2017.

\bibitem{Strub2017}
F.~Strub, H.~{De Vries}, J.~Mary, B.~Piot, A.~Courvile, and O.~Pietquin,
  ``{End-to-end optimization of goal-driven and visually grounded dialogue
  systems},'' \emph{IJCAI International Joint Conference on Artificial
  Intelligence}, pp. 2765--2771, 2017.

\bibitem{Bowman2016}
S.~R. Bowman, L.~Vilnis, O.~Vinyals, A.~Dai, R.~Jozefowicz, and S.~Bengio,
  ``Generating sentences from a continuous space,'' in \emph{Proceedings of The
  20th SIGNLL Conference on Computational Natural Language Learning}.\hskip 1em
  plus 0.5em minus 0.4em\relax Association for Computational Linguistics, 2016,
  pp. 10--21.

\bibitem{Kingma2014}
D.~P. Kingma and M.~Welling, ``Auto-encoding variational bayes.'' in
  \emph{International Conference on Learning Representations (ICLR)}, 2014.

\bibitem{Kusner2017}
M.~J. Kusner, B.~Paige, and J.~M. Hern{\'a}ndez-Lobato, ``Grammar variational
  autoencoder,'' in \emph{Proceedings of the 34th International Conference on
  Machine Learning}, ser. Proceedings of Machine Learning Research, D.~Precup
  and Y.~W. Teh, Eds., vol.~70.\hskip 1em plus 0.5em minus 0.4em\relax
  International Convention Centre, Sydney, Australia: PMLR, 06--11 Aug 2017,
  pp. 1945--1954.

\bibitem{Bahuleyan2017}
H.~Bahuleyan, L.~Mou, O.~Vechtomova, and P.~Poupart, ``Variational attention
  for sequence-to-sequence models,'' in \emph{Proceedings of the 27th
  International Conference on Computational Linguistics}.\hskip 1em plus 0.5em
  minus 0.4em\relax Association for Computational Linguistics, 2018, pp.
  1672--1682.

\bibitem{Song2016}
S.~Song, F.~Yu, A.~Zeng, A.~X. Chang, M.~Savva, and T.~Funkhouser, ``Semantic
  scene completion from a single depth image,'' \emph{IEEE Conference on
  Computer Vision and Pattern Recognition}, 2017.

\bibitem{Rissanen1976}
J.~J. Rissanen, ``Generalized kraft inequality and arithmetic coding,''
  \emph{IBM Journal of Research and Development}, vol.~20, no.~3, pp. 198--203,
  May 1976.

\bibitem{Rissanen1979}
J.~Rissanen and G.~G. Langdon, ``Arithmetic coding,'' \emph{IBM Journal of
  Research and Development}, vol.~23, no.~2, pp. 149--162, March 1979.

\bibitem{Witten1987}
I.~H. Witten, R.~M. Neal, and J.~G. Cleary, ``Arithmetic coding for data
  compression,'' \emph{Commun. ACM}, vol.~30, no.~6, pp. 520--540, Jun. 1987.

\bibitem{Shannon1948}
C.~Shannon, ``{A Mathematical Theory of Communication},'' \emph{Bell System
  Technology}, vol.~27, no.~I, pp. 379--656, 1948.

\bibitem{Cho2014}
K.~Cho, B.~van Merrienboer, C.~Gulcehre, D.~Bahdanau, F.~Bougares, H.~Schwenk,
  and Y.~Bengio, ``Learning phrase representations using rnn encoder--decoder
  for statistical machine translation,'' in \emph{Proceedings of the 2014
  Conference on Empirical Methods in Natural Language Processing
  (EMNLP)}.\hskip 1em plus 0.5em minus 0.4em\relax Association for
  Computational Linguistics, 2014, pp. 1724--1734.

\bibitem{Pascanu2014}
R.~Pascanu, {\c C}.~G{\"{u}}l{\c c}ehre, K.~Cho, and Y.~Bengio, ``How to
  construct deep recurrent neural networks,'' in \emph{International Conference
  on Learning Representations (ICLR)}, 2014.

\bibitem{Miller1995}
G.~A. Miller, ``Wordnet: A lexical database for english,'' \emph{Commun. ACM},
  vol.~38, no.~11, pp. 39--41, Nov. 1995.

\bibitem{Hochreiter1997}
S.~Hochreiter and J.~Schmidhuber, ``Long short-term memory,'' \emph{Neural
  Computation}, vol.~9, no.~8, pp. 1735--1780, 1997.

\bibitem{Li2018}
S.~Li, W.~Li, C.~Cook, C.~Zhu, and Y.~Gao, ``Independently recurrent neural
  network (indrnn): Building {A} longer and deeper {RNN},'' in \emph{Conference
  on Computer Vision and Pattern Recognition (CPVR)}, 2018.

\bibitem{Kingma2015}
D.~P. Kingma and J.~Ba, ``Adam: {A} method for stochastic optimization,'' in
  \emph{International Conference on Learning Representations (ICLR)}, 2015.

\bibitem{Glorot2010}
X.~Glorot and Y.~Bengio, ``Understanding the difficulty of training deep
  feedforward neural networks,'' in \emph{JMLR W\&CP: Proceedings of the
  Thirteenth International Conference on Artificial Intelligence and Statistics
  (AISTATS 2010)}, vol.~9, May 2010, pp. 249--256.

\bibitem{Saxe2014}
A.~M. Saxe, J.~L. McClelland, and S.~Ganguli, ``Exact solutions to the
  nonlinear dynamics of learning in deep linear neural networks,'' in
  \emph{International Conference on Learning Representations (ICLR)}, 2014.

\end{thebibliography}


\clearpage
\begin{appendices}

\section{Context-free grammar (CFG) used in the experiments}
\label{sec:appendix_cfg}


\begin{minipage}{\linewidth}
\scriptsize
\begin{lstlisting}
s -> q

q -> qword adjective ',' adjective 'and' adjective '?'
q -> qword adjective 'and' adjective '?'
q -> qword adjective '?'
q -> qword 'made' 'of' noun_material '?'
q -> qword preposition np '?'
q -> qword np '?'
q -> 'can' 'it' 'make' 'a' 'sound' '?'
q -> 'can' 'it' 'play' 'music' '?'
q -> 'can' 'it' 'speak' '?'

np -> determiner adjective adjective adjective noun
np -> determiner adjective ',' adjective 'and' adjective noun
np -> determiner adjective 'and' adjective noun 'made' 'of' noun_material
np -> determiner adjective adjective noun
np -> determiner adjective 'and' adjective noun
np -> determiner adjective noun 'made' 'of' noun_material
np -> determiner noun 'made' 'of' noun_material
np -> determiner adjective noun
np -> determiner noun

qword -> 'is' 'it' | 'is' 'the' 'object' | 'is' 'the' 'thing'
noun -> noun_object | noun_material | noun_roomtype
preposition -> preposition_spatial | preposition_spatial_rel | preposition_material

adjective -> adjective_color | adjective_affordance | adjective_overall_size | 
             adjective_relative_size | adjective_relative_per_dimension_size | 
             adjective_mass | adjective_state | adjective_other

noun_object -> 'accordion' | 'acoustic' 'gramophone' | 'bar' | 'barrier' |
               'basket' | 'outdoor' 'lamp' | 'outdoor' 'seating' | ...

noun_material -> 'bricks' | 'carpet' | 'decoration' 'stone' | 'facing' 'stone' | 
                 'grass' | 'ground' | 'laminate' | 'leather' | 'wood' | ...

noun_roomtype -> 'aeration' | 'balcony' | 'bathroom' | 'bedroom' | 'boiler' 'room' | 
                 'garage' | 'guest' 'room' | 'hall' | 'hallway' | 'kitchen' | ...

determiner -> 'a' | 'an' | 'that' | 'the' | 'this'

preposition_spatial -> 'behind' | 'in' 'front' | 'near' | 
                       'on' 'the' 'left' | 'on' 'the' 'right'
preposition_spatial_rel -> 'behind' 'of' | 'in' | 'in' 'front' 'of' | 
                           'near' | 'on' 'the' 'left' 'of' | 'on' 'the' 'right' 'of'
preposition_material -> 'made' 'of'

adjective_color -> 'antique' 'white' | 'magenta' | 'maroon' | 
                   'slate' 'gray' | 'white' | 'yellow' | ...
                   
adjective_affordance -> 'actable' | 'addable' | 'addressable' | 'deliverable' | 
                        'destroyable' | 'dividable' | 'movable' | ...
                        
adjective_size -> adjective_overall_size | adjective_relative_size |
                  adjective_relative_per_dimension_size
                  
adjective_overall_size -> 'average-sized' | 'huge' | 'large' | 'small' | 'tiny'
adjective_relative_size -> 'average-sized' | 'huge' | 'large' | 'small' | 'tiny'
adjective_relative_per_dimension_size -> 'deep' | 'narrow' | 'shallow' | 
                                         'short' | 'tall' | 'wide'

adjective_mass -> 'heavy' | 'light' | 'moderately' 'heavy' | 'moderately' 'light' | 
                  'slightly' 'heavy' | 'very' 'heavy' | 'very' 'light'
adjective_state -> 'closed' | 'opened'
adjective_other -> 'textured' | 'transparent'
\end{lstlisting}
\end{minipage}

\begin{table}
\begin{tabular}{|l|r|l|}
\hline
\multicolumn{1}{|c|}{\textbf{Annotation}} & \multicolumn{1}{|c|}{\textbf{Nb. of classes}} & \multicolumn{1}{|c|}{\textbf{Example of classes}} \\ \hline
SUNCG category & 86 & air conditioner, mirror, window, door, piano \\ \hline
WordNet category & 580 & instrument, living thing, furniture, decoration \\ \hline
Location & 24 & kitchen, bedroom, bathroom, office, hallway, garage \\ \hline
Color & 139 & red, royal blue, dark gray, sea shell \\ \hline
Color property & 2 & transparent, textured \\ \hline
Material & 15 & wood, textile, leather, carpet, decoration stone \\ \hline
Overall mass & 7 & light, moderately light, heavy, very heavy \\ \hline
Overall size & 4 & tiny, small, large, huge \\ \hline
Category-relative size & 10 & tiny, small, large, huge, short, shallow, narrow, wide \\ \hline
State & 2 & opened, closed \\ \hline
Acoustical capability & 3 & sound, speech, music \\ \hline
Affordance & 100 & attach, bend, divide, play, shake, stretch, wear \\ \hline
\end{tabular} \vspace{1ex}
\caption{Description of all annotations that can be automatically derived from the SUNCG dataset \cite{Song2016} and other sources (e.g. WordNet \cite{Miller1995}). The category annotations derived from SUNCG and WordNet describe the type of the objects. From the 3D models in SUNCG, multiple colors and materials (based on textures) can be associated with the objects. The overall mass and size classes are computed according to all objects (i.e. a table is heavier and bigger than a book). The category-specific sizes are computed relative to objects in the same category (i.e. a specific table may be smaller and wider than another table model). The annotations also includes information about the state of the objects (e.g. is a door closed or opened), and the acoustical capability (e.g. can it produce sound, or music). An extensive list of affordances (e.g. can the object be moved or cleaned) is also provided.}
\label{tb:annotations}
\end{table}

\clearpage
\subsection{Size of the language space}

From the CFG used in the experiment, it is possible to extract a total of 15,396 distinct grammar rules. as shown below. In the case of the unbiased dataset, those rules can produce a total of 9.81e+18 unique sentences. While it is impractical to compute, the total number of unique sentences for the biased dataset is expected to be an order of magnitude smaller.

\begin{minipage}{\linewidth}
\scriptsize
\begin{lstlisting}
[qword, prep_material, determiner, adj_state, 'and', adj_other, noun_roomtype, '?']
[qword, prep_spatial, determiner, adj_other, adj_state, adj_state, noun_object, '?']
[qword, determiner, adj_other, ',', adj_mass, 'and', adj_affordance, noun_roomtype, '?']
[qword, determiner, adj_relative_per_dimension_size, adj_overall_size, noun_object, '?']
[qword, determiner, adj_overall_size, ',', adj_state, 'and', adj_state, noun_material, '?']
[qword, prep_spatial, determiner, adj_other, adj_mass, adj_affordance, noun_material, '?']
[qword, adj_state, 'and', adj_relative_size, '?']
[qword, prep_material, determiner, adj_mass, adj_other, adj_other, noun_material, '?']
[qword, prep_spatial, determiner, adj_state, adj_other, adj_color, noun_object, '?']
[qword, determiner, adj_relative_size, 'and', adj_overall_size, noun_material, '?']
[qword, determiner, adj_state, adj_overall_size, adj_other, noun_roomtype, '?']
[qword, determiner, adj_other, adj_state, adj_mass, noun_material, '?']
[qword, determiner, adj_overall_size, 'and', adj_other, noun_material, '?']
[qword, determiner, adj_color, adj_other, noun_object, '?']
[qword, prep_spatial_rel, determiner, adj_mass, adj_color, noun_roomtype, '?']
[qword, determiner, adj_state, 'and', adj_relative_size, noun_object, '?']
[qword, determiner, adj_color, adj_color, adj_relative_size, noun_material, '?']
[qword, determiner, adj_affordance, noun_object, '?']
[qword, determiner, adj_other, adj_other, adj_state, noun_roomtype, '?']
\end{lstlisting}
\end{minipage}

\clearpage
\section{Example of sentences generated from the CFG}

\subsection{Biased dataset}

\begin{minipage}{\linewidth}
\small
\begin{lstlisting}
is it the transparent door ?
is it a small toy ?
is it reachable and transferable ?
is it cultivatable , shallow and substitutable ?
is it small and graspable ?
is it gray and heavy ?
is it the alice blue and beige chandelier ?
is it this powder blue light cyan tall refrigerator ?
is it textured , average-sized and saddle brown ?
is it light gray and deep ?
is it a graspable large dining table ?
is it a short , misty rose and floral white kitchen cabinet ?
is it movable , small and silver ?
is it that textured indian red picture frame ?
\end{lstlisting}
\end{minipage}

\subsection{Unbiased dataset}

\begin{minipage}{\linewidth}
\small
\begin{lstlisting}
is the object this shelf made of grass ?
is the thing in front of that surveillance camera ?
is it a yellow range hood ?
is the object a toilet ?
is it in front of the peru armchair ?
is the object near a pale golden rod measurable wireless telephone ?
is it the sea green and pale golden rod air conditioning made of wallpaper ?
is it extendable , shrinkable and large ?
is it on the right a salmon carpeting made of bricks ?
is it a physical body made of stone ?
\end{lstlisting}
\end{minipage}

\section{Continuous sentence embedding using arithmetic coding}
\label{sec:appendix_arithmetic_coding}

The multidimensional extension of arithmetic coding is as follows: if the arithmetic coder is allowed to successively split into intervals an embedded space of $N_d$ dimensions, then it simply rotates among the dimensions as symbols are processed in the sequence. This means the first symbol in the sequence will lead to interval splits over first dimension, the second symbol over the second dimension, and so forth. If the length of the sequence $N_s$ is larger than $N_d$, then the dimension $n$ used at each iteration $i \in \lbrace 1, 2, \dots, N_s \rbrace$ will be $n_i = i \; mod \; N_d$. If $N_s$ is much smaller than $N_d$, then some dimensions will never be used. To avoid this, one can multiply the output vector by a random orthonormal matrix to cover all dimensions. The decoder only needs to apply the inverse transform before the actual decoding.

\begin{figure}[h!]
\centering
\includegraphics[width=1.0\linewidth]{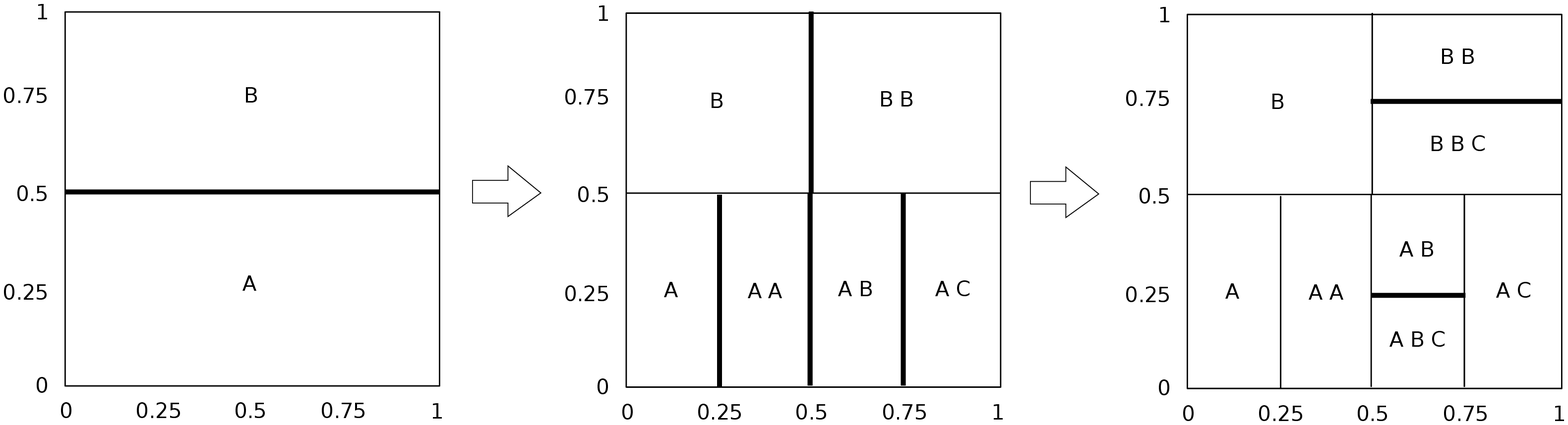}
\cprotect\caption{Continuous sentence embedding using arithmetic coding. In this example where $N_d = 2$, the encoded CFG is  \verb!S -> A | B | A A | A B | A C | B B | A B C | B B C! .}
\label{fig:grammar_arithmetic_coding_splits}
\end{figure}

\section{Continuous sentence embedding using recurrent neural networks (RNNs)}
\label{sec:appendix_rnn}

We performed the experiments with GRU \cite{Cho2014} units for all methods as they have fewer parameters to learn than the LSTM. Furthermore, we did not get different results with LSTM \cite{Hochreiter1997} and IndRNN \cite{Li2018} units during preliminary evaluations.

For all RNN-based embeddings, we used the Adam \cite{Kingma2015} optimizer with a learning rate of 1e-3 and gradient clipping at 0.5 magnitude. During training, the learning was reduced by a factor of 0.2 if the loss function didn't decrease in the last 5 epochs, but with a minimum learning rate of 1e-5. Kernel weights used the Xavier uniform initialization \cite{Glorot2010}, while recurrent weights used random orthogonal matrix initialization \cite{Saxe2014}. All biases were initialized to zero.

\end{appendices}


\end{document}